\documentclass[10pt,twocolumn,letterpaper]{article}

\usepackage{wacv}              

\usepackage{graphicx}
\usepackage{amsmath}
\usepackage{amssymb}
\usepackage{booktabs}

\usepackage[pagebackref,breaklinks,colorlinks]{hyperref}

\usepackage[capitalize]{cleveref}
\crefname{section}{Sec.}{Secs.}
\Crefname{section}{Section}{Sections}
\Crefname{table}{Table}{Tables}
\crefname{table}{Tab.}{Tabs.}

\def\wacvPaperID{11} 
\def\confName{WACV}
\def\confYear{2025}

\begin{document}

\title{SIGN: A Statistically-Informed Gaze Network for Gaze Time Prediction}

\author{Jianping Ye \qquad Michel Wedel\\ 
The University of Maryland\\
College Park, MD 20742, USA\\
{\tt\small \{jpye00,mwedel\}@umd.edu}
}
\maketitle

\begin{abstract}
We propose a first version of SIGN, a Statistically-Informed Gaze Network, to  predict aggregate gaze times on images. We develop a foundational statistical model for which we derive a deep learning implementation involving CNNs and Visual Transformers, which enables the prediction of overall gaze times. The model enables us to derive from the aggregate gaze times the underlying gaze pattern as a probability map over all regions in the image, where each region’s probability represents the likelihood of being gazed at across all possible scan-paths. We test SIGN's performance on AdGaze3500, a dataset of images of ads with aggregate gaze times, and on COCO-Search18, a dataset with individual-level fixation patterns collected during search. We demonstrate that SIGN (1) improves gaze duration prediction significantly over state-of-the-art deep learning benchmarks on both datasets, and (2) can deliver plausible gaze patterns that correspond to empirical fixation patterns in COCO-search18. These results suggest that the first version of SIGN holds promise for gaze-time predictions and deserves further development. 
\end{abstract}

\section{Introduction}
\label{sec: Intro}




Building human-like visual systems is a fundamental goal in machine learning. The recognition of the central role of human eye movements in visual perception and attention 
has led to a surge of interest in the development of deep learning models of human gaze. For example \cite{kummerer2014deep, kummerer2022deepgaze} show that Convolutional Neural Networks (CNN) can generate human-like scanpaths better than traditional methods, such as \cite{itti1998model}. 
Long Short-Term Memory (LSTM)  models 
\cite{cornia2018predicting, chen2021predicting}, inverse reinforcement learning \cite{chen2021coco}, Transformers \cite{mondal2023gazeformer} and Large Language Models \cite{chen2025gazexplain}, have enabled the sequential modeling of gaze patterns by including high-level semantic information to improve visual tasks such as object detection \cite{liu2016dhsnet, wang2018salient}, semantic segmentation \cite{zeng2019joint ,lee2021railroad}, visual question answering \cite{chen2025gazexplain}, and decision making \cite{byrne2023predicting, unger2024predicting}.

For training and inference, these prior models rely on eye-tracking data comprising of individual-level fixation patterns. Instead, in the present study we develop a computational gaze model that is trained on \textit{aggregate gaze time data}. Aggregate gaze data arise frequently when data collection is constrained by data privacy laws (such as GDPR) or because of the costs, complexity and scalability of data storage and processing pipelines. Research companies collect many hundreds of eye-tracking datasets per year, and oftentimes collect or retain only the aggregated gaze times on commercial images. 
Aggregate gaze times are of great interest because they reflect the depth of attention and offer insights into cognitive engagement in applications such as advertising and human-computer interaction \cite{pieters2004attention, zhang2009sales}. 

Therefore, we develop a machine learning model to predict gaze times, called  
SIGN (Statistically-Informed Gaze Network). We start with a foundational statistical model that extends prior work (e.g., \cite{shi2013information, liechty2003global}), which we  implement as a deep neural network that enables prediction of aggregate gaze time data. SIGN allows us to derive from the aggregate gaze times the underlying gaze pattern as a probability map over all spatial regions in the image, where each region’s probability represents the likelihood of attracting gaze across all possible scan-paths. We validate our model on two datasets: AdGaze3500 and COCO-Search18; the former contains aggregate gaze times on advertisements, the latter individual-level fixation sequences on images of scenes. The results show that our approach achieves better performance in gaze time prediction than state-of-the-art deep learning methods, while the inferred gaze patterns align well with empirical fixation patterns.

\section{Methods}
We first propose a foundational statistical scan-path model. 
Our contribution over prior statistical models 
\cite{shi2013information,liechty2003global} is that we relax the first order Markov assumption and include image features. The model describes aggregate gaze times, for which previously mostly linear, hierarchical linear, and tree models have been used \cite{pieters2004attention,zhang2009sales, ye2024contextual}. The model assumes scan-paths, trajectories of fixations of the eyes \cite{noton1971scanpaths}, to underlie gaze time on images that appear within a context (such as images on websites, or ads within magazines). 
\label{sec: Methods}
\subsection{Gaze Model Formulation}
The foundational model is based on the following assumptions, which are grounded in prior empirical findings.
\paragraph{Assumption 1.} Gaze time is \textit{additive} in the log space of time, that is, the log gaze time arises as a sum of a series of log fixation durations on regions of the image. 
\paragraph{Assumption 2.} The log gaze $g$ for an image is generated by the following stochastic process with the state space $\mathcal{S}=\{S_1,...,S_N\}$, consisting of local $K$-dimensional feature vectors for $N$ regions on image $I$: let $(F_t)_{t=1}^{\Tilde{T}}\in\{1,\dots,N\}^{\Tilde{T}}$ be fixations on regions of $I$, $I^c$ be the context if present, with $\Tilde{T}\in\mathbb{Z}^+$ the (predetermined) total number of fixations. For $t \leq \Tilde{T}$ the generative process is:
\begin{enumerate}
    \item During scene exploration the first fixation often lands in the center \cite{tatler2007central}; within a single fixation the gist of the scene is extracted \cite{oliva2006building}. Generate $F_0$ to be the center point of the blurred and down-scaled image $\check{I}$, with features $S_0$ that capture the ``gist". Assuming that the influence of the context operates via its gist, generate the initial fixation duration $g_0=\mu_0(S_0,S_c)$, with features $S_c$ that capture the gist of the context $I^c$.  
    \item According to Inhibition of Return (IoR) \cite{klein2000inhibition} visited states are less likely to be visited again. Generate the transition probability $\hat{E}$ from previous fixation $F_{t-1}$ to the next state $S_j$ with $\hat{E}_j=\hat{E}_j((S_{F_{t-k}})_{k=1}^{t-1}, S_j)$, such that $\sum_j \hat{E}_{j}=1$ and $\hat{E}_{F_l}=0$ for all $l=1,\dots,t$. Sample the next fixation $F_{t}$ from $\text{Categorical}(\hat{E})$.
    \item Fixation durations are affected by image features \cite{nuthmann2017fixation}. Generate log-fixation duration $g_t$, $t\geq1$, from some distribution $\mathcal{D}$ with mean $\mu(S_{F_{t}})$ governed by features at the current fixation on the image, $S_{F_{t}}$. 
\end{enumerate}
\paragraph{Remarks:} (1) Because the length of a scan-path is finite,  we assume it is bounded above by $\Tilde{T}$; (2) The (infinite horizon) IoR can be relaxed to a fixed time horizon.
\paragraph{Assumption 3.} If $G$ is the gaze time on the image $I$ with context $I^c$, then:
$$\mathbb{E}(G) = \exp\left({\mathbb{E}g}\right),$$
where $g=g_0+\sum_{t=1}^{\Tilde{T}}g_t$ generated from the process in Assumption 2. This assumption allows us to use observed aggregate gaze times to fit the model. 

\paragraph{Problem Formulation.} Here we formulate our model based on the three assumptions. We divide $I$ into equal spatial regions (image patches) with non-overlapping local features (states) $\{S_1,\dots,S_{N}\}$. We use $\mu_0$ to represent the initial log-fixation duration on the features $S_0$ of the blurred image $\check{I}$, and $\mu$ to represent the log-fixation duration determined by local features $S_j$ of  region $j$. Since the length of a scan-path is finite, the expected log gaze $g$ is:
\begin{equation}
\label{eqn:gaze_for_average}
\mathbb{E}(g) = \mu_0(S_0,S_c) + \sum_{\tau}P_{\tau}\sum_i \mu\left(S_{\tau_i}\right),
\end{equation}
where $\tau$ is one scan-path $(F_t)_{t=1}^{\Tilde{T}}$, $i$ indicates the regions visited in the scan-path $\tau$, and $P_{\tau}$ is the associated probability of $\tau$. Note that $P_{\tau}$ is a function of the features $\{S_j\}$, since the transition probabilities are determined by local features of the image by step 2 of assumption 2. Because we assume that we have no access to individual fixation locations, we invoke IoR (Assumption 2) and rearrange the terms, which yields:
\begin{equation}
\label{eqn:gaze_for_average_equivalent}
\mathbb{E}(g) = \mu_0(S_0,S_c) + \sum_{j=1}^N\mu(S_j)w_j,
\end{equation}
with weights $w_j$ defined by
\begin{equation}
\label{eqn:weights}
    w_j = \sum_{\tau\in \mathcal{T}_{S_j}}P_{\tau},
\end{equation}
where $\mathcal{T}_{S_j}$ is the set of all scan-paths that contain region $S_j$. Let us call the first and the second terms in equation \ref{eqn:gaze_for_average_equivalent} ``gist" gaze and total local gaze, respectively. As a result of Assumption 3, equation \ref{eqn:gaze_for_average_equivalent} can be estimated if  one replaces the left hand side by the log of the empirical average gaze time. This is important in many practical situations where only the average gaze time data is available. 

\paragraph{Inferred Gaze Pattern} We define the inferred fixation pattern as:
\begin{equation}
\label{eqn: gaze_pattern}
    p_{S_j} = \frac{w_j}{\sum_k w_k}=\frac{\sum_{\tau\in \mathcal{T}_{S_j}}P_{\tau}}{\sum_{S_k}\sum_{\tau\in \mathcal{T}_{S_k}}P_{\tau}}
\end{equation}
for $j=1,\dots,N$. We observe that the weights in equation \ref{eqn:weights} indicate the frequency with which an individual will fixate on local features $S_j$: the larger the weight is, the more likely $S_j$ will attract gaze. Normalizing the weights of all regions generates a probability map that represents the fixation pattern over the regions, integrated across all possible scan-paths.

\subsection{SIGN Architecture}

We develop the SIGN deep learning architecture to estimate the complex fixation duration ($\mu_0$ and $\mu$) and weight functions ($w_j$) in equations~\ref{eqn:gaze_for_average_equivalent} and \ref{eqn:weights}. Appendix 1 contains a diagram of the model's architecture.

\paragraph{Modeling the Fixation Duration Functions $\mu_0$ and $\mu$.} We use CNNs \cite{lecun1998gradient} to extract visual features, and use Multi-Layer Perceptrons (MLPs) to model the fixation-duration functions in relation to the image features. Particularly, $\mu_0$ is produced by a MLP taking features $S_0$ and $S_c$ from fine-tuneable ResNet50 models \cite{he2016deep} on blurred $\check{I}$, respectively $I^c$, and $\mu$ is another MLP taking local features $S_j$ from a simple fresh 5-layer CNN, whose inputs are raw image patches (regions).

\paragraph{Modeling the Weights $w_j$.} The weights in equation \ref{eqn:weights} are a function of all local features $S_j$. The Transformer \cite{vaswani2017attention} is a natural candidate for the spatial modeling of those weights, as it integrates all visual features on the image. 
The Transformer takes in local features $S_j$ from a simple 5-layer CNN. A MLP is then applied to learn a single weight for each transformed feature. A sigmoid activation is applied to the MLP output to ensure $w_j \in (0,1)$ (see equation~\ref{eqn:gaze_for_average_equivalent}). Hence, the weights are modeled as:

\begin{equation}
    w_j = \sigma\left(\text{MLP}\left(\text{Trf}(\left[S_1,\dots,S_N\right])_{j}\right)\right),
\end{equation}
where $\text{Trf}(\cdot)$ is the Transformer, $S_j$ is the $j$th local feature vector, $\text{Trf}(\cdot)_{j}$ denotes the $j$th ``transformed" local feature vector, and $\sigma(\cdot)$ is the sigmoid activation.

\paragraph{Context Effects} If context $I^c$ is present we include its effects in $\mu_0$ (see the foundational model). The same fine-tuneable ResNet50 model used for the image extracts the visual features $S_c$ from a down-scaled and blurred $I^c$, which are then concatenated with $S_0$, those of the blurred image $I$, and fed into a MLP which produces $\mu_0$ from the concatenated feature vectors.

\section{Preliminary Results}
\label{sec: prelim_results}

We apply SIGN to two datasets, and compare its results with several benchmarks, in terms of accuracy of gaze prediction. 

\subsection{Data}
\paragraph{AdGaze3500 \cite{ye2024contextual}} is a commercial dataset that contains 3531 digital display advertisements from the Dutch market. These ads come from 71 categories and appeared in 29 different mainstream digital magazines. Each ad image is either double-page or single-page accompanied by a counter-page (the context). Participants were asked to read the magazines freely as they would in their daily life. For each ad aggregate gaze data is available, collected by eye tracking of around 80-100 regular customers. Individual-level fixation locations are not available, making the training of SIGN more challenging. 

\paragraph{COCO-Search18 \cite{chen2021coco}} is a large-scale dataset that involves eye movements of ``Target Absent" (TA) search tasks on 2489 images. Each image was shown to 10 participants and their eye fixations were recorded. Because we are interested in assessing the effectiveness of SIGN to recover personalized fixation patterns, our experiments use the gaze time data from the TA task of all images for only the first participant. We focus on the data of one individual to compare SIGN's predicted weight maps with that individual's observed fixations, but because the data is sparse we do not expect SIGN to be highly accurate in many cases and the results are reported here as an illustration only. 

\subsection{Training Procedure and Hyperparameters} Both datasets are randomly split into 90\% for training and 10\% for testing. 10-fold cross-validation is applied to the training data for an ensemble of 10 models. Images in the AdGaze3500 dataset were resized to $256 \times 256$ (height by width), and those in COCO-Search18 were resized to $256 \times 416$. We use vanilla ResNet and Transformer models. The CNNs used to model local features $S_j$ have 1 layer of $5\times5$ and 4 layers of $3 \times 3$ convolutions. We trained the free parameters of SIGN with respect to Mean Square Error (MSE) loss between the model predictions and the ground-truth; the parameters of its ResNet50 and Transformer modules are fine-tuned; all model parameters are trained simultaneously. For COCO-Search18, because the fixations are sparse we add a weak sparsity constraint on the weights in equation \ref{eqn:weights} as prior knowledge. The model is optimized by Adam optimizer \cite{kingma2014adam}. We start the optimization with learning rates $1e-3$ and $1e-4$ for the AdGaze3500 and COCO-Search18 datasets respectively. We half the learning rate every 5 epochs with 60 epochs in total. We patchify the images as was done in \cite{dosovitskiy2020image} to obtain the spatial regions. 

\subsection{Patch Size Affects SIGN Performance}
\label{subsec:patch_size}
We test patch sizes of $8 \times 8$, $16 \times 16$, and $32 \times 32$ pixels on the AdGaze3500 dataset to evaluate their impact on prediction performance. We hypothesize that an optimal patch size exists that maximizes predictive accuracy: small regions may distort high-level semantic information, while large regions may miss fine-grained local details. 
Test-data loss ($8 \times 8$: 0.137, $16 \times 16$: 0.134, and $32 \times 32$: 0.135) reveals that a patch size of $16 \times 16$ is the optimal choice. An ablation experiment with AdGaze3500 shows that the gist of the context substantially reduces test-data loss (0.134 vs. 0.179).


\subsection{SIGN Presents Better Gaze Predictions}
We compare the performance of the proposed SIGN model with state-of-the-art baselines, including ResNet50 and ResNet101 \cite{he2016deep}, a Vision Transformer (ViT) \cite{dosovitskiy2020image} 
, and a DeepGaze Salience-based \cite{kummerer2014deep} MLP fine-tuned on each dataset. We use test-data MSE and the correlation between the model predictions and the ground truth as performance measures. Table \ref{tab:gaze_prediction_accuracy} presents the results. We observe that the proposed SIGN model outperforms all baselines on both datasets and on both measures. First, the improvement obtained by adding in SIGN a local gaze module to the ResNet50 model clearly indicates SIGN's ability to extract more fine-grained local information that facilitates gaze prediction. Such an improvement is not simply obtained by introducing more parameters, because SIGN outperforms ResNet101 even though it uses fewer parameters ($\sim37$M) than ResNet101 ($\sim43$M). Apparently it is the architecture of SIGN rooted in our foundational model that produces improved prediction accuracy. Further we find that compared with the ViT, with a powerful spatial attention mechanism, ($\sim87$M) parameters, and like SIGN $16 \times 16$ local image patches, SIGN achieves much higher predictive accuracy. Finally, comparing SIGN with DeepGaze shows SIGN's contribution over a state-of-the-art salience model. This shows that SIGN more efficiently exploits local features than a wide range of established models.

\begin{table}[!htb]
    \centering
    \begin{tabular}{l|cc|cc}
    \hline
    & \multicolumn{2}{c|}{AdGaze3500} & \multicolumn{2}{c}{COCO-Search18} \\
    \hline
     &  MSE $\downarrow$ &  Corr. $\uparrow$ &  MSE $\downarrow$ &  Corr. $\uparrow$ \\
    \hline
    ResNet50 &  0.141    &  0.81  & 0.216 & 0.61\\
    ResNet101 & 0.138    &  0.82  & 0.221 & 0.60\\
    ViT       & 0.194    &  0.63  & 0.218 & 0.60\\
    DeepGaze+MLP & 0.192 & 0.68 & 0.232 & 0.50\\
    \hline
    SIGN      & \textbf{0.134}    & \textbf{0.83} & \textbf{0.212} & \textbf{0.62}\\
    \hline
    \end{tabular}
    \caption{Comparing prediction accuracy across different models with MSE loss and correlation (Corr.) on the test data of the two datasets. Each result is the average of all 10 models from the 10-fold cross-validation.}
    \label{tab:gaze_prediction_accuracy}
\end{table}

\subsection{SIGN Delivers Plausible Inferred Gaze Patterns}
\begin{figure}[!htb]
    \centering
    \includegraphics[width=1.0\linewidth]{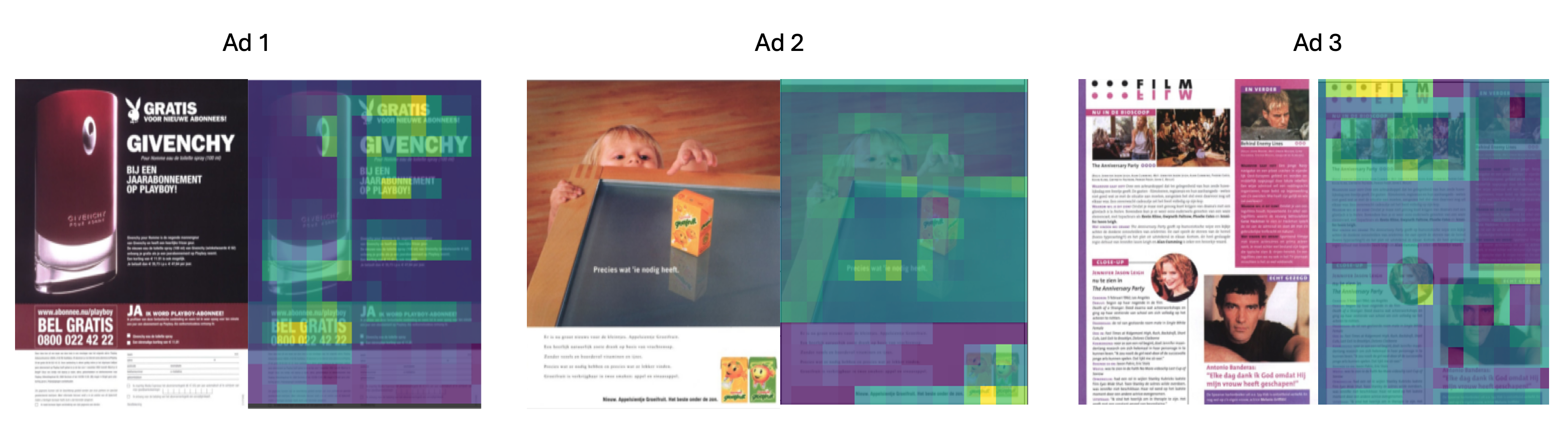}
    \caption{Sample predicted weight maps for AdGaze3500. For each sample, the left half presents the original image, and the right half presents the image overlaid with the generated SIGN weights. Brighter regions are predicted to be more conspicuous.}
    \label{fig:ad_maps}
\end{figure}

\begin{figure}[!htb]
    \centering
    \includegraphics[width=1.0\linewidth]{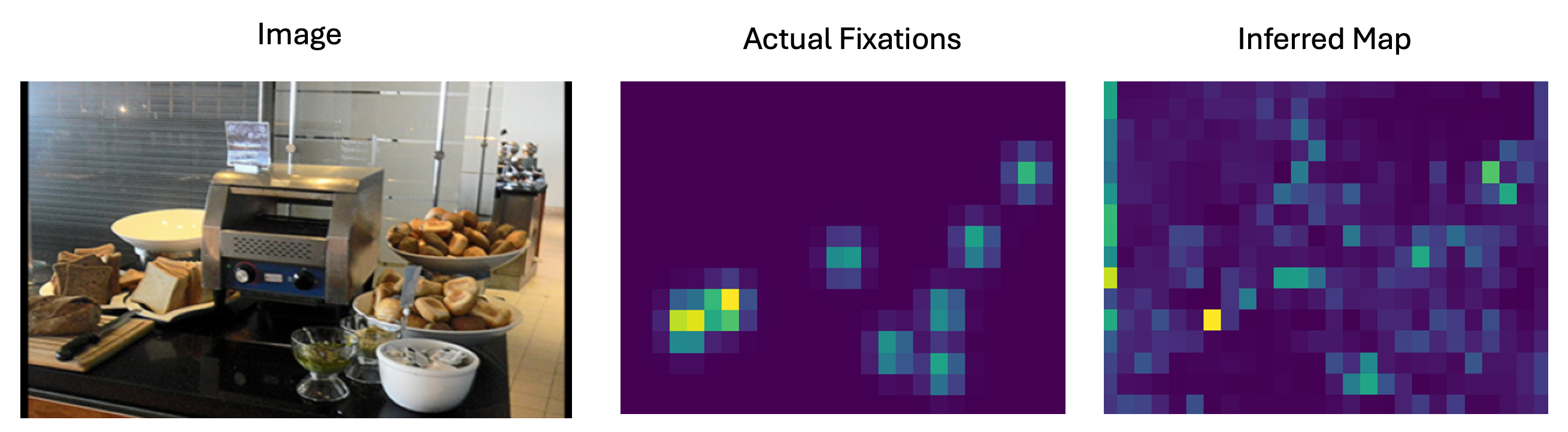}
    \caption{Sample predicted weight map for COCO-Search18 data. The original image is on the left, the actual fixation locations in the middle (blurred by a Gaussian filter with a standard deviation of 35 pixels \cite{bylinskii2018different} and resized), and the predicted inferred map is on the right.}
    \label{fig:search_maps}
\end{figure}

We normalized the trained weights in Equation \ref{eqn: gaze_pattern} and plotted the resulting probability maps for the test images. Figures \ref{fig:ad_maps} and \ref{fig:search_maps} show a few (non cherry-picked) samples; Appendix 2a and 2b contain additional samples. It is clear that the predicted maps trained on the AdGaze3500 data accurately capture headlines, products, and text blocks, which are regarded as the key design-regions of ads that attract gaze \cite{pieters2004attention}. Due to the unstructured nature of the images from the COCO-Search18 data, the gaze pattern inferred from the probability maps is less well explainable. Nonetheless, for this indivdual's data the predicted map overlaps with the observed fixations to a large extent, illustrating that the inferred map is plausible, probabilistically.

\section{Conclusion and Future Work}
\label{sec:conclusion}
SIGN, our neural network model implementation of a foundational statistical formulation rooted in theories of attention and eye movements, shows promise in (1) predicting gaze time on images, and (2) generating plausible gaze patterns even when no individual fixation data are available. 

While the initial results show that the generated gaze patterns correspond with observed ones, even when the model is trained on only aggregate gaze data, the accuracy of the SIGN-generated gaze patterns needs to be verified more extensively. Currently, it is not yet clear if SIGN generates accurate individual-level gaze maps for all individual search tasks in the COCO-Search18 data. The reasons are that first, the gaze pattern for individual-level search tasks are extremely sparse, and second that the images in this dataset have very heterogeneous spatial structures which makes it difficult for SIGN to generalize. Finally, currently SIGN is restricted to equally sized spatial regions, while people tend to look at regions that are semantically coherent and irregular hypothesize that SIGN may perform better for regions that contain coherent semantic information. 

Overall, these initial findings provide valuable insights into  SIGN's strengths and areas that may require further optimization through subsequent experimentation and model refinement. In addition, SIGN can be used as a reward model to improve gaze on advertisements, with the ultimate goal of better matching advertised products and services to consumers' needs, and thus have a positive societal impact. 


{\small
\bibliographystyle{ieee_fullname}
\bibliography{egbib}
}
\newpage

\begin{figure*}[!t]
    \centering
    \captionsetup{labelformat=empty}
    \includegraphics[width=0.9\linewidth]{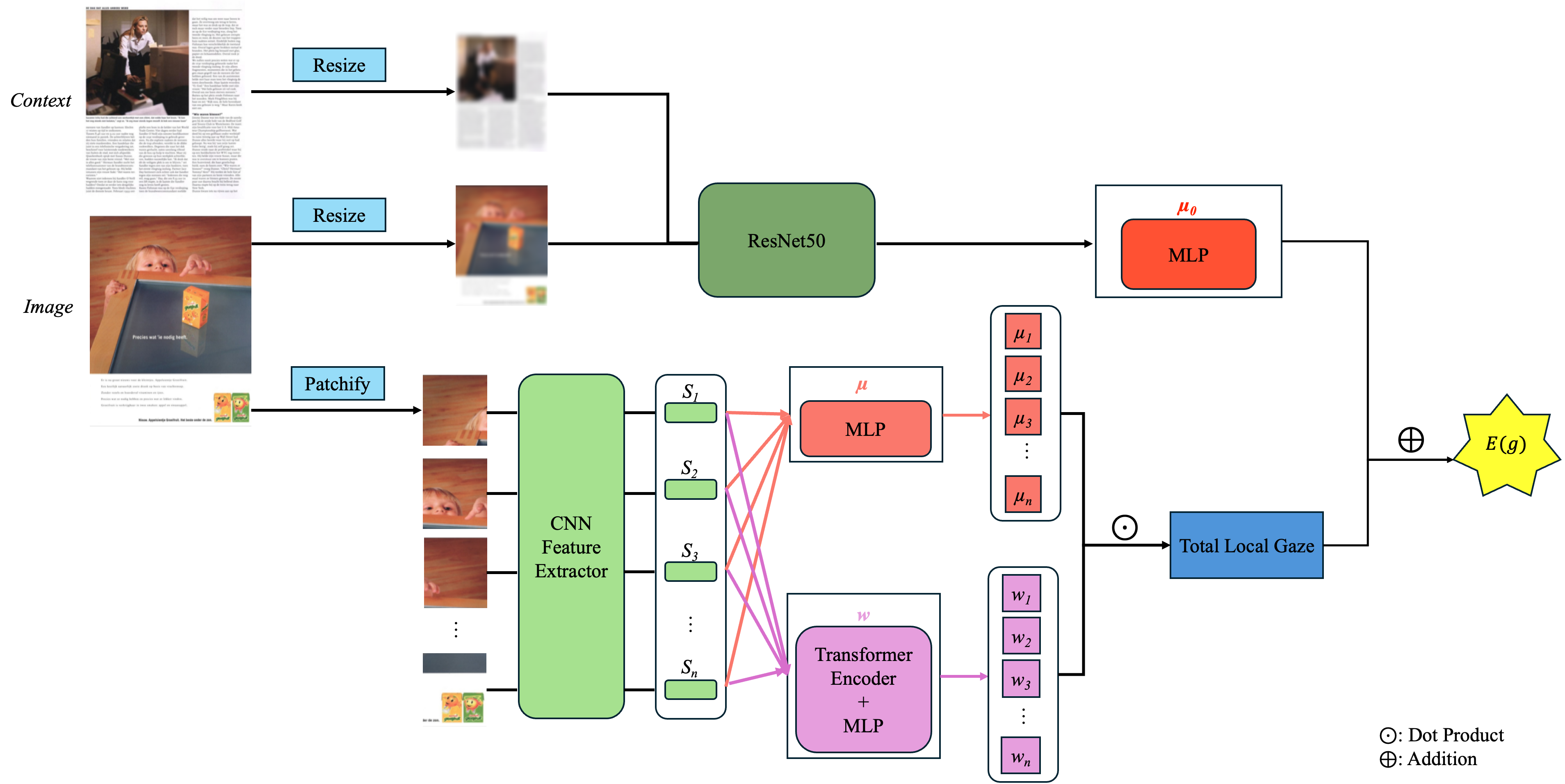}
    \caption{\textbf{Appendix 1: Diagram of SIGN Architecture}. The architecture consists of two separate parts, respectively corresponding to image and context `gist' $\mu_0$ and total  gaze on image regions defined in equation \ref{eqn:gaze_for_average_equivalent}. The ``gist" is represented by the global visual features extracted by a pre-trained and fine-tunable ResNet50 block, which are then passed to $\mu_0$, modeled by an MLP, to calculate the gaze induced by the ``gist". In parallel, the original image is patchified into regions, which are subsequently compressed to local visual features $S_1,S_2,\dots,S_n$ generated by a trainable CNN. Each local feature $S_i$ is used to determine local gaze $\mu_i$ through an MLP; local features are also used to calculate local weights $w_i$, calculated from another MLP on top of a transformer encoder that enhances interactions among local features. The total local gaze pattern is calculated as the dot product between the local gaze and their local weights; the final predicted gaze is the sum of the ``gist" gaze and the total local gaze pattern.}
    \label{fig:SIGN_Diagram}
\end{figure*}

\begin{figure*}[!htb]
    \centering
    \captionsetup{labelformat=empty}
    \includegraphics[width=0.8\linewidth]{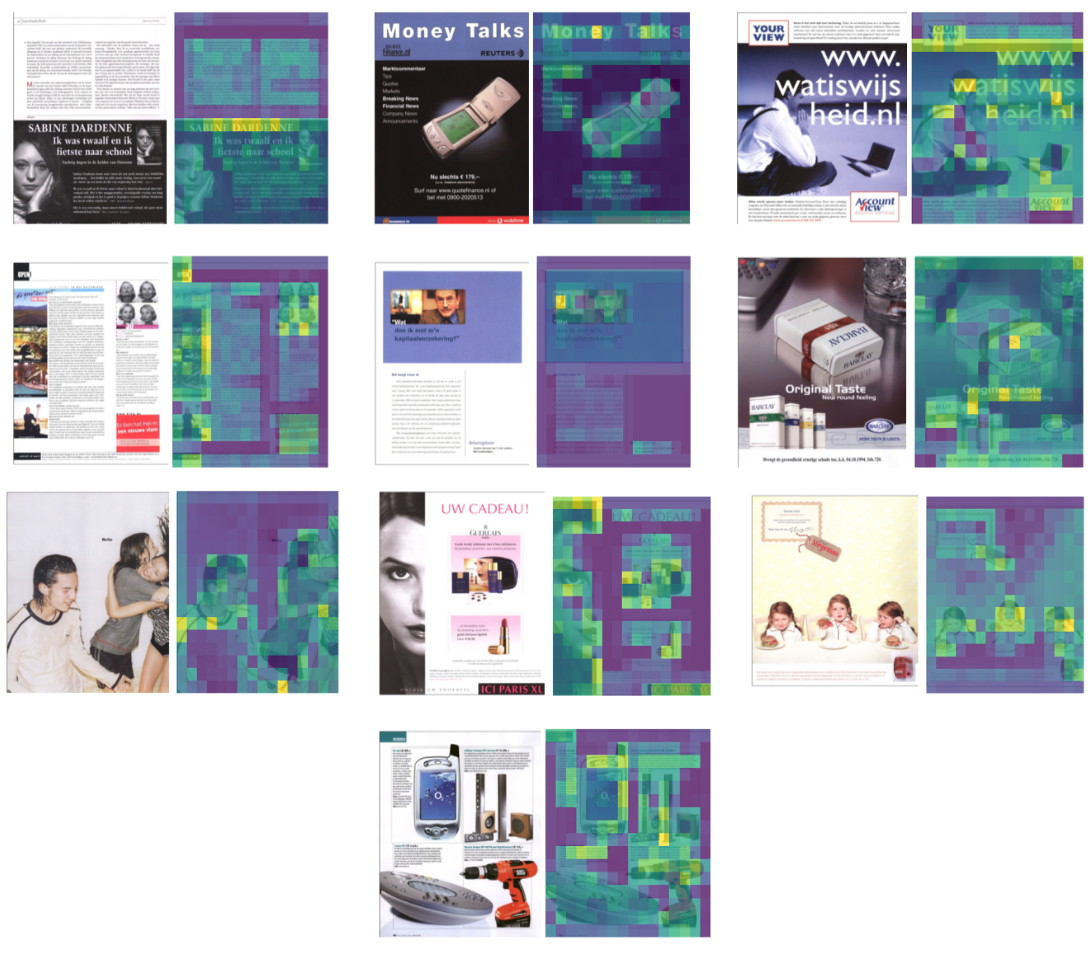}
    \caption{\textbf{Appendix 2a: Additional Predicted Weight Maps for AdGaze3500.}}
    \label{fig:extra_weight1}
\end{figure*}

\begin{figure*}[!htb]
    \centering
    \captionsetup{labelformat=empty}
    \includegraphics[width=0.8\linewidth]{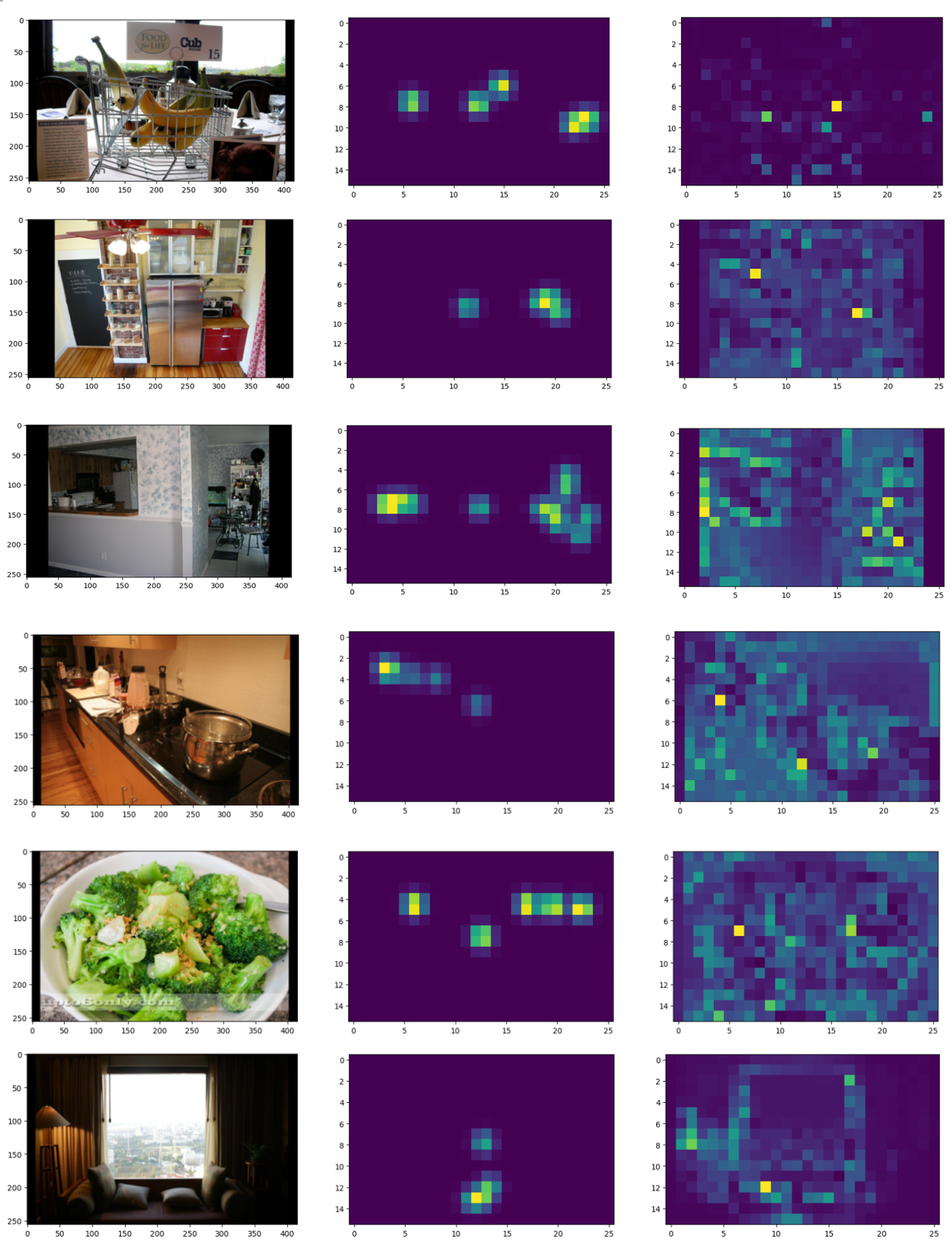}
    \caption{\textbf{Appendix 2b: Additional Predicted Weight Maps for COCO-Search 18.}}
    \label{fig:extra_weight2}
\end{figure*}

\end{document}